%% file: main.tex
\documentclass[letterpaper]{article} 
\usepackage{aaai2026}  
\usepackage{times}  
\usepackage{helvet}  
\usepackage{courier}  
\usepackage[hyphens]{url}  
\usepackage{graphicx} 
\usepackage{natbib}  
\usepackage{caption} 
\usepackage{algorithm}
\usepackage{algorithmic}

\usepackage{newfloat}
\usepackage{listings}

\usepackage{amsmath}
\usepackage{multirow}
\usepackage{booktabs}
\usepackage{url}
\usepackage{subcaption}
\usepackage[misc]{ifsym}
\usepackage[table]{xcolor}
\usepackage[export]{adjustbox}
\usepackage{array}

\newcommand{\classifier}{%
    \includegraphics[height=1em, valign=m]{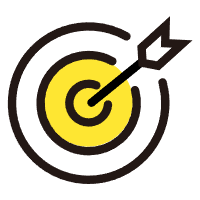}%
}
\newcommand{\generative}{%
    \includegraphics[height=1em, valign=m]{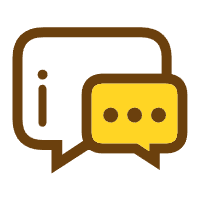}%
}

\DeclareCaptionStyle{ruled}{labelfont=normalfont,labelsep=colon,strut=off} 
\floatstyle{ruled}
\newfloat{listing}{tb}{lst}{}
\floatname{listing}{Listing}
\begin{document}
%
\title{Long-form RewardBench: Evaluating Reward Models for Long-form Generation}

\author{
    Hui Huang\textsuperscript{1}$^{*}$, 
    Yancheng He\textsuperscript{1}$^{*}$, 
    Wei Liu\textsuperscript{1}$^{*}$,
    Muyun Yang\textsuperscript{1}$^\dagger$,
    Jiaheng Liu\textsuperscript{2}$^\dagger$, \\
    \Large\textbf{Kehai Chen\textsuperscript{3},}
    \textbf{Bing Xu\textsuperscript{1},}
    \textbf{Conghui Zhu\textsuperscript{1},}
    \textbf{Hailong Cao\textsuperscript{1},}
    \textbf{Tiejun Zhao\textsuperscript{1}}    
}

\affiliations{
    \textsuperscript{\rm 1}Faculty of Computing, Harbin Institute of Technology, Harbin, China \\
    \textsuperscript{\rm 2}School of Intelligence Science and Technology, Nanjing University, Suzhou, China \\
    \textsuperscript{\rm 3}School of Computer Science and Technology, Harbin Institute of Technolgy, Shenzhen, China \\
    \text{huanghui@stu.hit.edu.cn, yangmuyun@hit.edu.cn}

}


\maketitle

\let\oldthefootnote\thefootnote
\let\thefootnote\relax\footnotetext{$*$ Equal contribution.}
\let\thefootnote\relax\footnotetext{$\dagger$\:Corresponding Author.}
\let\thefootnote\oldthefootnote

\input{content/0_Abstract}

\begin{links}
    \link{Code}{https://github.com/HuihuiChyan/Long-formRMB}
\end{links}

\input{content/1_Introduction}
\input{content/2_Background}
\input{content/3_Methodology}
\input{content/4_Experiments}
\input{content/5_Analysis}
\input{content/6_Conclusion}

\bibliography{aaai2026}

\end{document}

%% file: content/0_Abstract.tex
\begin{abstract}
The widespread adoption of reinforcement learning-based alignment highlights the growing importance of reward models. Various benchmarks have been built to evaluate reward models in various domains and scenarios. However, a significant gap remains in assessing reward models for long-form generation, despite its critical role in real-world applications. To bridge this, we introduce \textbf{Long-form RewardBench}, the first reward modeling testbed specifically designed for long-form generation. Our benchmark encompasses five key subtasks: QA, RAG, Chat, Writing, and Reasoning. We collected instruction and preference data through a meticulously designed multi-stage data collection process, and conducted extensive experiments on 20+ mainstream reward models, including both classifiers and generative models. Our findings reveal that current models still lack long-form reward modeling capabilities. Furthermore, we designed a novel \textbf{Long-form Needle-in-a-Haystack Test}, which revealed a correlation between reward modeling performance and the error's position within a response, as well as the overall response length, with distinct characteristics observed between classification and generative models. Finally, we demonstrate that classifiers exhibit better generalizability compared to generative models trained on the same data. As the first benchmark for long-form reward modeling, this work aims to offer a robust platform for visualizing progress in this crucial area\textsuperscript{1}.
\end{abstract}

%% file: content/1_Introduction.tex
\section{Introduction}

Reward Models are designed to simulate human preferences to enhance the training effectiveness of language models. Typically, reward models learn from preference data and output a scalar value proportional to the quality of the input text \cite{zhong2025comprehensive}. Reward models have been widely applied in Reinforcement Learning from Human Feedback (RLHF) training and also play crucial roles in direct alignment algorithms, data filtering, and inference-time scaling \cite{he2025can,huang2025think,zhang2024prototypical}.

Although the research community has developed several benchmarks to evaluate the preference modeling capabilities of reward models \cite{lambert2025rewardbench,malik2025rewardbench}, existing evaluation texts are relatively short, typically containing only tens to hundreds of tokens. However, long-form text generation presents many unique challenges, such as textual coherence, information consistency, and overall structural integrity \cite{que2024hellobench}. The long-form problem necessitates a specially designed reward model benchmark, for the purpose of driving progress in generating high-quality long texts that align with human expectations.

Therefore, in this study, we introduce \textbf{Long-form RewardBench}, a novel benchmark designed to comprehensively evaluate the preference modeling capabilities of various long-form reward models. It encompasses five key areas: \textit{Question Answering (QA)}, \textit{Retrieval Augmented Generation (RAG)}, \textit{Chat}, \textit{Writing}, and \textit{Reasoning}. To ensure the benchmark's robustness, we meticulously collected instructions and responses from carefully selected datasets and representative models. Preference annotation was then achieved through a multi-stage LLM-as-a-Judge process. Furthermore, to simulate real-world applications of reward models, we employed the \textit{best-of-n classification} evaluation method as our primary means of assessment.

We evaluated over 20 representative reward models on Long-form RewardBench, which are categorized as two types: \textit{sequence classifier} and \textit{generative model}. Our evaluation revealed a significant performance gap in their ability to model the effectiveness of long-form text generation compared to general text generation. This underscores the need for more targeted design and optimization of reward models specifically for the long-form domain.

Moreover, we designed a novel \textbf{Long-form Needle-in-a-Haystack Test}, to investigate the correlation between long-form reward modeling accuracy with error position, as well as response length. Our findings indicate that the accuracy of generative models show a high correlation with both error position and response length, whereas sequence classifiers demonstrate robustness to response length variations. Furthermore, we conducted experiments investigating the correlation between training data length and reward modeling accuracy, and reveal that generative models are notably more sensitive to training sequence length, whereas sequence classifiers generalize more effectively.

Our main contributions are as follows:

\begin{enumerate}
    \item We introduce Long-form RewardBench, the first reward modeling benchmark for long-form generation, by meticulously designed preference construction process.
    \item We evaluated 20+ reward models on Long-form RewardBench, uncovering the current limitations in their long-form reward modeling capabilities.
    \item We developed the Long-form Needle-in-a-Haystack Test, revealing that while error position significantly impacts accuracy, classifier models demonstrate better generalizability regarding sequence length.
\end{enumerate}


%% file: content/2_Background.tex
\section{Background}
\subsection{Long-form Generation}
The increasing use of LLMs in professional settings demands ever-longer responses. Recent progress in long-form text generation has enabled models to produce outputs thousands of words long. Various benchmarks have also been established to evaluate LLMs' long-form generation capabilities \cite{que2024hellobench,wu2024longgenbench}, revealing that current long-output LLMs still struggle with issues like local incoherence and internal contradictions.

Previous efforts to extend long-form generation typically relied on supervised fine-tuning using synthetic datasets \cite{wu2025longwriter,pham2024surimulticonstraintinstructionfollowing}. However, given the scarcity of high-quality annotated data and the inefficiencies of imitation learning, recent research is shifting towards reinforcement learning-based methods to enhance long-form generation \cite{wu2025longwriter}, which underscores the critical need for robust long-form reward modeling.

\subsection{Reward Modeling Evaluation}
\label{sec:background2}

Reward models play a central role in incorporating human preferences into LLMs, especially in alignment techniques like Reinforcement Learning from Human Feedback (RLHF) \cite{zhong2025comprehensive,dong2024rlhfworkflowrewardmodeling}. They receive prompts and responses as input, then score or rank these responses based on human preferences (e.g., helpfulness, harmlessness, and factuality). The generated reward signal is then used to supervise the optimization of LLMs.

Common types of reward models include:

\begin{enumerate}
    \item \textbf{Sequence Classifier} This method appends a regression head to the model to predict a reward score.
    \item \textbf{Generative Model} This method uses a generative model to produce reward scores or rankings directly as text.
\end{enumerate}

Early reward model evaluations focused on simple classification tasks to measure the performance of existing reward models in common aspects like style and safety \cite{lambert2025rewardbench}. Subsequent evaluations have included analyzing reward models' inference methods (such as best-of-n sampling) and their downstream scores when trained with RLHF \cite{malik2025rewardbench,frick2024evaluate}. Some work also focuses on specific aspects of reward evaluation, such as fine-grained evaluation \cite{liu2024rm}, multimodal evaluation \cite{yasunaga2025multimodalrewardbenchholisticevaluation}, multilingual evaluation \cite{gureja2024m}, math reasoning \cite{kim2024evaluating} and RAG evaluation \cite{jin2024rag}. 

However, existing reward benchmarks still concentrate on responses of tens to hundreds of tokens. With the prevalence of long-form generation, there is an urgent need for a reward benchmark specifically designed for long-form scenarios.

%% file: content/3_Methodology.tex
\section{Long-form RewardBench}

\begin{figure*}[!t]
    \centering
    \includegraphics[width=\textwidth]{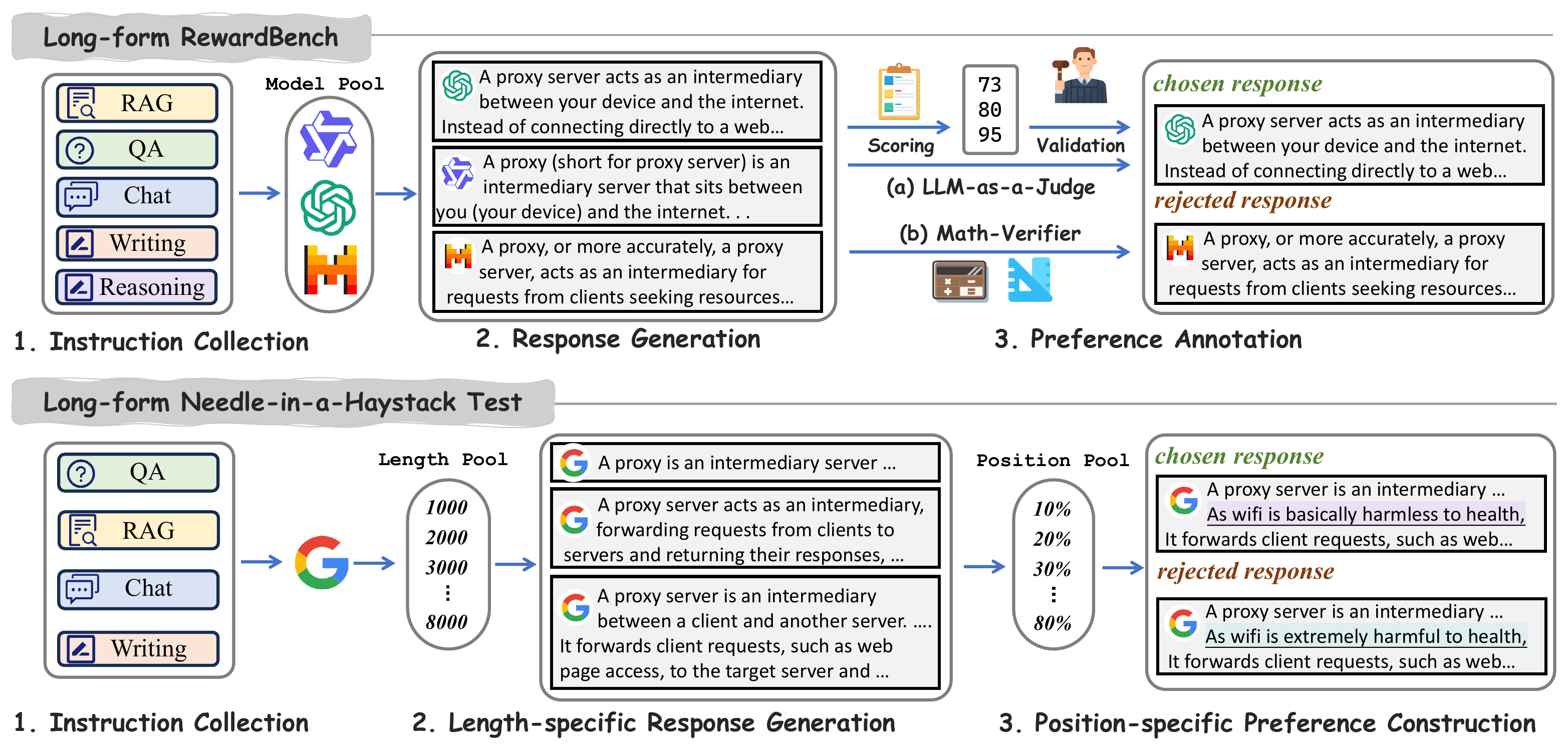} 
    \caption{The data construction process of Long-form RewardBench and Long-form Needle-in-a-Haystack Test.}
    \label{fig:length-dist}
\end{figure*}

\subsection{Benchmark Construction}

\subsubsection{Instruction Collection}

To ensure the diversity and representativeness of the benchmark, we carefully screened and sampled long-form queries from multiple existing datasets, covering five representative application scenarios:

\begin{itemize}

    \item \textbf{Question Answering (QA)}: This category targets the ability to provide comprehensive and informative answers to complex questions, especially in the open-ended scenario. We sampled queries with an output length exceeding 500 tokens from the QuoraQA \footnote{\url{toughdata/quora-question-answer-dataset}}.
    \item \textbf{Retrieval-Augmented Generation (RAG)}: This category evaluates the model's capacity to generate accurate and relevant long-form text based on retrieved information. We sampled queries with an output length exceeding 1000 tokens from RAGBench \cite{jin2024rag}.
    \item \textbf{Chat}: This scenario focuses on open-ended conversational queries, mimicking real-world interactive dialogues where detailed and extensive responses are often required. We sampled queries with an output length exceeding 2000 tokens from WildChat \cite{zhao2024wildchat}.
    \item \textbf{Writing}: This scenario assesses the generation of creative and extended written pieces, such as articles, stories, or reports. We sampled 500 queries from LongWriter-6k \cite{bai2024longwriter}.
    \item \textbf{Reasoning}:  This scenario challenges the model's problem-solving abilities across math domain. We selected 1000 queries from DeltaBench \cite{he2025can}, which includes multiple subsets such as Omni-MATH \cite{gao2024omnimathuniversalolympiadlevel}, OlympiadBench \cite{he2024olympiadbench}, and MATH \cite{hendrycks2021measuring}.
\end{itemize}

We utilized gemini-2.5-pro to score all instructions from 1-10, and retained only those with a score above 7 to ensure that the instructions are of high clarity and effectiveness.

\subsubsection{Response Generation}

After obtaining the high-quality instructions, we leverage over 15 representative models to generate the responses, which can be categorized into two groups: open-sourced and close-sourced models. 

\begin{itemize}
    \item \textbf{Open-sourced Models} This includes Qwen-2.5 \cite{team2024qwen2}, Llama-3 \cite{grattafiori2024llama}, Deepseek-V3 \cite{liu2024deepseek}, Mistral \cite{jiang2023mistral7b} with different model sizes, all with instruction-finetuned versions.
    \item \textbf{Close-sourced Models} This includes gpt-4o, claude3.5, grok-3, gemini-1.5, gemini-2.0, gemini-2.5, etc.
\end{itemize}

We filtered out all responses shorter than 200 tokens, thereby guaranteeing that all responses in the final benchmark fall into the long-form category.

\subsubsection{Preference Annotation}

Preference annotation is a crucial step in constructing a high-quality reward model benchmark. In this work, we employed a multi-stage automated process, combined with a rule-based method, specifically tailored to the characteristics of different data types.

For the QA, RAG, Chat and Writing subsets, where subjective quality plays a significant role, we annotated preferences using a multi-stage LLM-as-a-Judge process\footnote{The prompt templates used for LLM-judge based preference annotation are presented in Appendix A.1.}:

\begin{enumerate}
    \item Criterion Generation: We leveraged gpt-4o to automatically generate specific scoring criteria for each instruction, along with corresponding weights that sum to 10.
    \item Pointwise Scoring: We utilized gpt-4o to score responses against these criteria, and then aggregated weighted individual scores into a holistic score.
    \item Preference Selection: We randomly designated one response as ``chosen" and selected three lower-scoring ``rejected" responses. If three lower-scoring responses weren't available, the chosen response was re-selected.
    \item Pairwise Validation: We used gemini-2.5-pro to validate each chosen response against its rejected counterparts. A pair was valid only if the chosen response unequivocally surpassed all three rejected responses.
\end{enumerate}

For the Reasoning subset, which has objectively verifiable outcomes, we used a direct rule-based method for preference annotation. We applied existing verification methods to determine correctness: responses with a correct solution were classified as chosen examples, while incorrect responses were classified as rejected.

\begin{table}[t]
\centering
\resizebox{\linewidth}{!}{
\begin{tabular}{lll|ccc}
\hline
\textbf{Subset}            & \textbf{Num}          & \textbf{Response} & \textbf{Avg.Len} & \textbf{Min.Len} & \textbf{Max. Len} \\ \hline
\multirow{2}{*}{QA}        & \multirow{2}{*}{227}  & Chosen            & 1463             & 327              & 4129              \\
                           &                       & Rejected          & 1013             & 258              & 3075              \\ \hline
\multirow{2}{*}{RAG}       & \multirow{2}{*}{239}  & Chosen            & 995              & 200              & 3332              \\
                           &                       & Rejected          & 640              & 200              & 6053              \\ \hline
\multirow{2}{*}{Chat}      & \multirow{2}{*}{332}  & Chosen            & 2286             & 569              & 8387              \\
                           &                       & Rejected          & 1719             & 200              & 32723             \\ \hline
\multirow{2}{*}{Writing}   & \multirow{2}{*}{328}  & Chosen            & 2779             & 520              & 10631             \\
                           &                       & Rejected          & 2090             & 200              & 32692             \\ \hline
\multirow{2}{*}{Reasoning} & \multirow{2}{*}{458}  & Chosen            & 6182             & 775              & 36545             \\
                           &                       & Rejected          & 5878             & 385              & 32677             \\ \hline
\multirow{2}{*}{Overall}   & \multirow{2}{*}{1816} & Chosen            & 3208             & 200              & 36545             \\
                           &                       & Rejected          & 2741             & 200              & 32723             \\ \hline
\end{tabular}}
\caption{Statistics of responses in different subsets.}
\label{tab:token_stats}
\end{table}

\subsubsection{Evaluation Method}

Given past research indicating a disconnect between reward model benchmark performance and actual effectiveness \cite{malik2025rewardbench,zhou2024rmb}, we opted for Best-of-N (BoN) sampling to evaluate our reward models. A sample is deemed as correct only when the chosen response is unequivocally superior to the rejected ones, allowing for accuracy calculation.

We calculate accuracy for each subset individually. The overall accuracy, used for benchmark ranking, is then determined by a weighted average across all categories.

For the Reasoning subset, due to the challenge of obtaining three incorrect responses for every correct one, we instead use pairwise accuracy as our evaluation metric.

\begin{figure}[!t]
    \begin{subfigure}[b]{0.48\textwidth}
        \hspace*{-0.1in}
        \includegraphics[width=\textwidth]{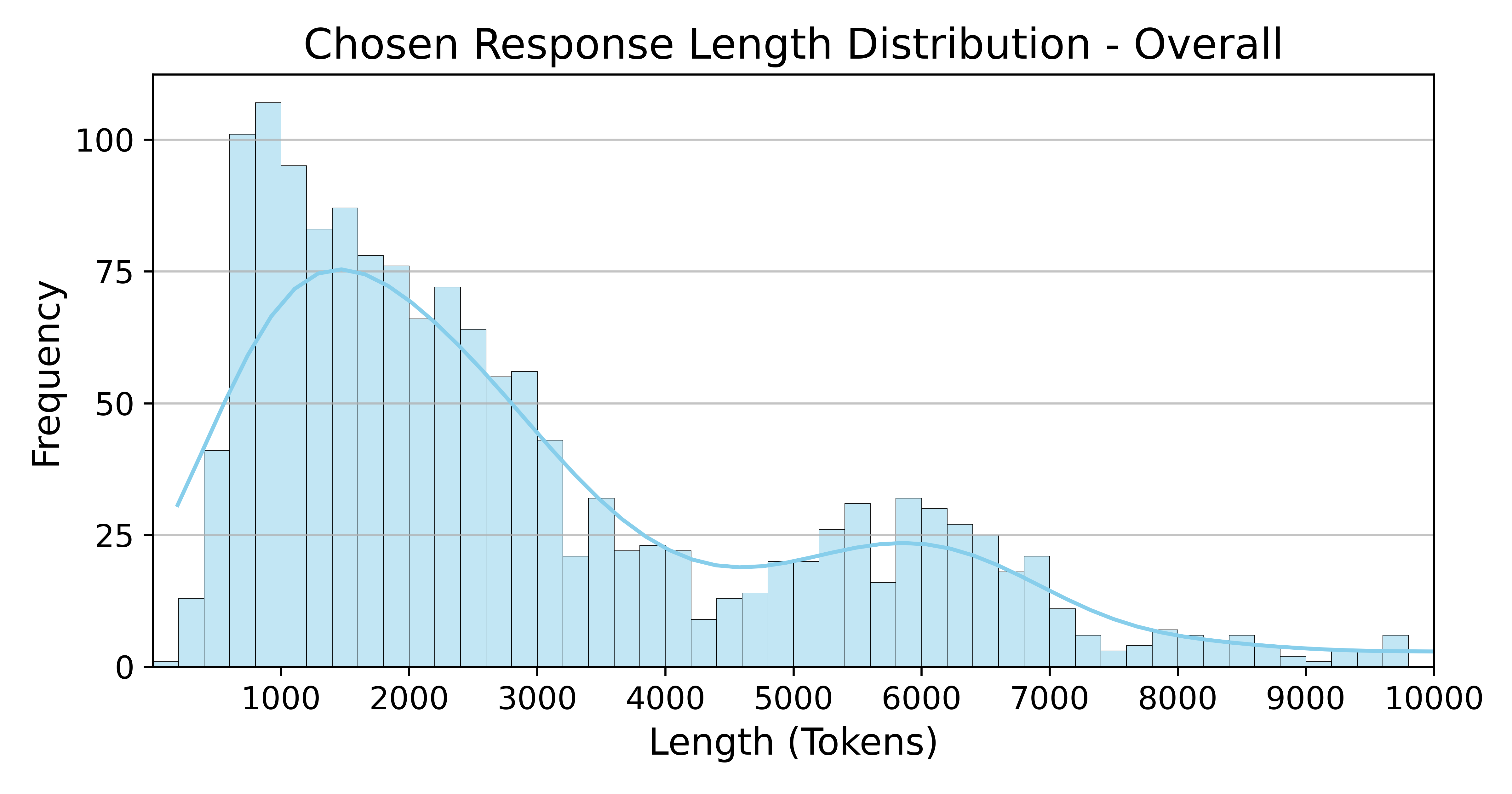} 
        \label{fig:subim1}
    \end{subfigure}
    \begin{subfigure}[b]{0.48\textwidth}
        \hspace*{-0.1in}
        \includegraphics[width=\textwidth]{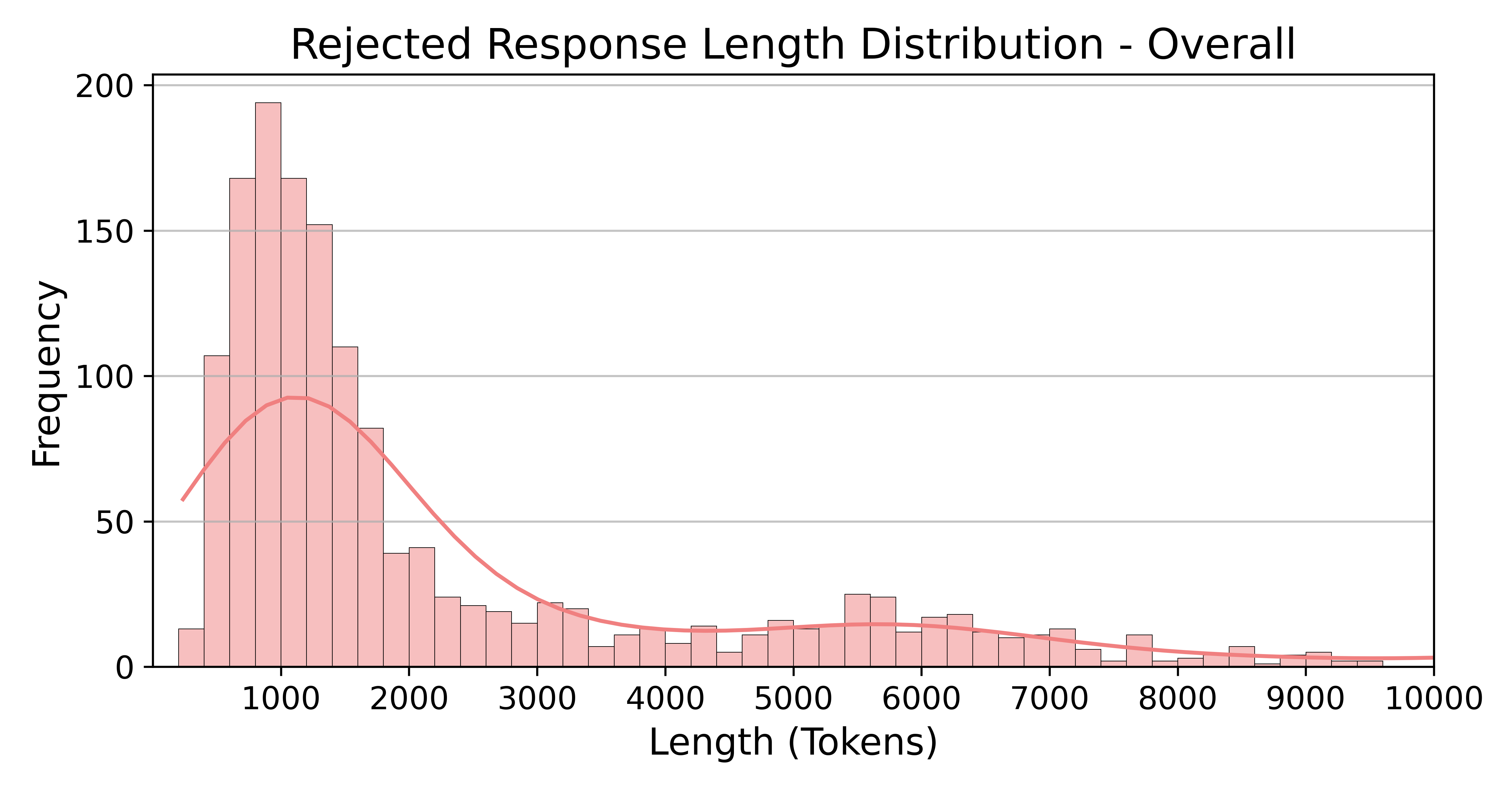} 
        \label{fig:subim2}
    \end{subfigure}
    \caption{Response length distribution of chosen and rejected responses on Long-form RewardBench.}
    \label{fig:length-dist}
\end{figure}

\subsection{Data Statistics}

This section presents the statistics of Long-form RewardBench. We first present the subset statistics in Table \ref{tab:token_stats}\footnote{Length is determined by \texttt{o200k\_base} from \texttt{tiktoken}.}, showing a balanced distribution of approximately 200-400 samples per category, striking a compromise between evaluation efficiency and comprehensiveness.

Figure \ref{fig:length-dist} illustrates the length distribution of responses\footnote{Distributions on all subsets are presented in Appendix A.2.}. While both chosen and rejected responses primarily fall within the 1K-2K token range, rejected texts exhibit a notably higher frequency in shorter lengths. However, this does not necessarily indicate a verbosity bias \cite{saito2023verbosity}, as detailedness is crucial in long-form scenarios, and more extensive responses often contain more useful information.

\begin{figure}[!t]
    \centering
    \begin{subfigure}[b]{0.48\textwidth}
        \centering
        \includegraphics[width=\textwidth]{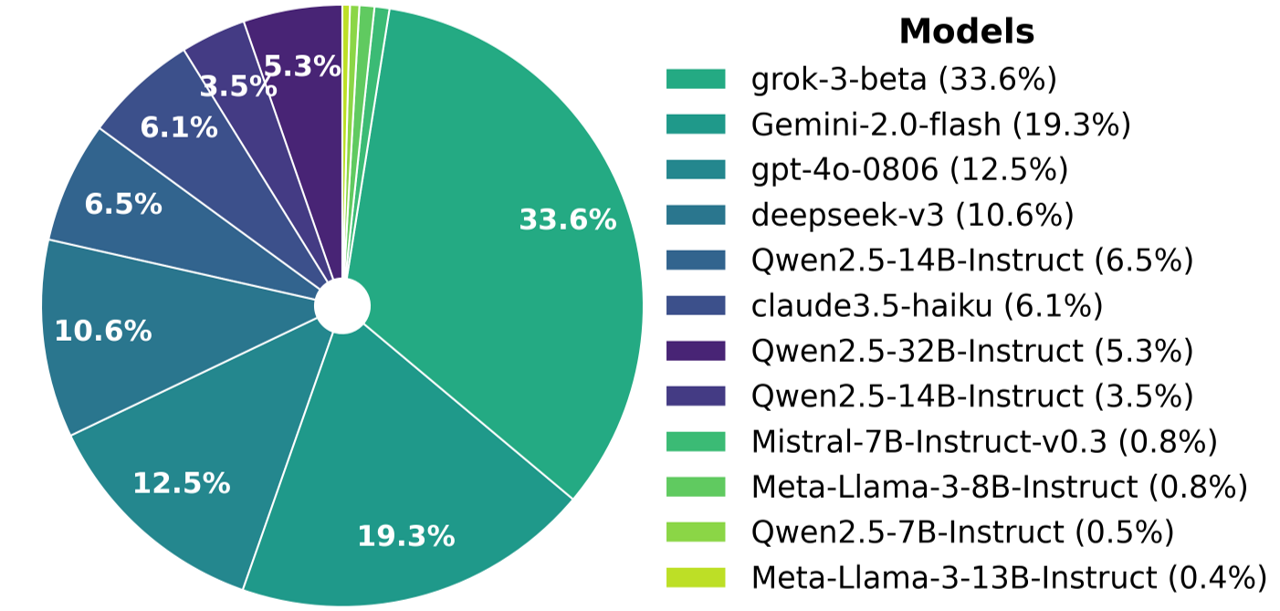} 
        \label{fig:subim1}
    \end{subfigure}
    \hfill 
    \begin{subfigure}[b]{0.48\textwidth}
        \centering
        \includegraphics[width=\textwidth]{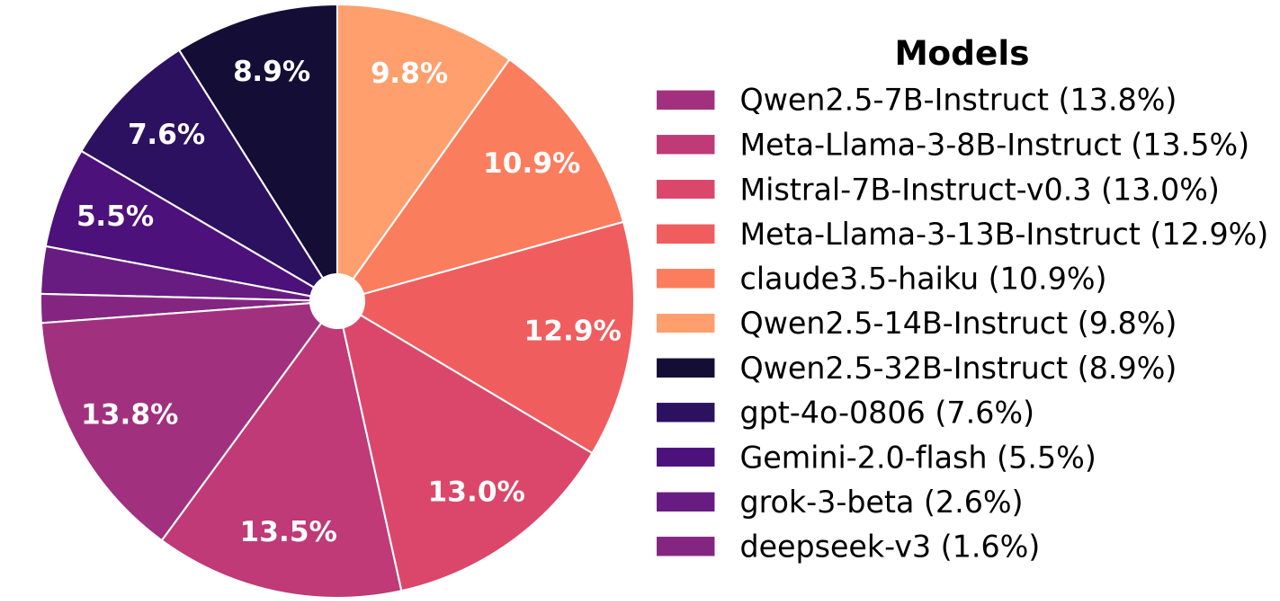} 
        \label{fig:subim2}
    \end{subfigure}
    \caption{Model distribution for chosen and rejected responses on Long-form RewardBench.}
    \label{fig:model-dist}
\end{figure}

Figure \ref{fig:model-dist} also presents the top 10 models for both chosen and rejected categories. As observed, a significant portion of the chosen responses originate from proprietary, closed-source models, which are generally believed to have undergone more extensive training and possess larger parameter sizes. Conversely, rejected responses predominantly come from smaller models. Furthermore, the distribution of rejected models is more even compared to that of chosen models. This is because each chosen response is required to pair with three inferior answers, which makes it challenging for comparably weaker models to generate chosen responses.

%% file: content/4_Experiments.tex
\subsection{Evaluation Set-up}
Based on the constructed Long-form RewardBench, we aim to conduct an in-detailed study of the recent reward models for their long-form reward modeling capabilities. Our evaluation is conducted on two groups of reward models, \textbf{sequence classifier} and \textbf{generative model}.


For sequence classifier, we directly leverage the model to assign a score to each chosen and rejected response. For generative models, they are evaluated under two settings\footnote{The prompt templates used for generative reward models for different settings are presented in Appendix A.1.}: 1) \textbf{Scoring}: Leverage the model to assign a score to each response; 2) \textbf{Selection}: Leverage the model to select the best response given all chosen and rejected responses. 

Moreover, apart from commonly used generative models such as gpt-4o, we also included fine-tuned generative reward models, such as RISE \cite{yu2025improvellmasajudgeabilitygeneral}, which is specifically fine-tuned for judgment.

\begin{table*}[t]
\resizebox{1.0\textwidth}{!}{

\begin{tabular}{cc|cccccc|c|c}
\hline
                                         &                                 & \multicolumn{6}{c|}{\textbf{Long-form-RM}}                                                           & \textbf{RB2}   & \textbf{RB1}   \\
\multirow{-2}{*}{\textbf{Model}}         & \multirow{-2}{*}{\textbf{Type}} & \textbf{Score} & \textbf{QA} & \textbf{RAG} & \textbf{Chat} & \textbf{Writing} & \textbf{Reasoning} & \textbf{Score} & \textbf{Score} \\ \hline
\rowcolor[HTML]{E0FFE0} 
Skywork/Skywork-Reward-V2-Llama-3.2-3B   & \raisebox{-1pt}{\classifier\hspace*{-0.15em}} Classifier                      & \textbf{74.9}  & 87.5        & 57.3         & 72.8          & 78.6             & 78.7               & 74.9           & -              \\
\rowcolor[HTML]{FFFFCC } 
openai/gpt-4.1-2025-04-14                & \raisebox{-1.2pt}{\generative\hspace*{-0.1em}} Generative                      & \textbf{74.6}  & 58.0        & 70.6         & 80.5          & 78.1             & 78.4               & 72.3           & -              \\
\rowcolor[HTML]{FFFFCC } 
anthropic/claude-opus-4-20250514         & \raisebox{-1.2pt}{\generative\hspace*{-0.1em}} Generative                      & \textbf{74.4}  & 57.1        & 71.9         & 75.4          & 79.4             & 80.1               & 76.5           & -              \\
\rowcolor[HTML]{E0FFE0} 
allenai/Llama-3.1-70B-Instruct-RM-RB2    & \raisebox{-1pt}{\classifier\hspace*{-0.15em}} Classifier                      & \textbf{74.4}  & 59.7        & 77.5         & 78.3          & 81.6             & 72.3               & 76.1           & 90.2           \\
\rowcolor[HTML]{E0FFE0} 
allenai/Llama-3.1-Tulu-3-70B-SFT-RM-RB2  & \raisebox{-1pt}{\classifier\hspace*{-0.15em}} Classifier                      & \textbf{73.9}  & 59.7        & 74.0         & 81.2          & 83.1             & 69.2               & 72.2           & 88.9           \\
\rowcolor[HTML]{FFFFCC } 
google/gemini-2.5-flash-preview-04-17    & \raisebox{-1.2pt}{\generative\hspace*{-0.1em}} Generative                      & \textbf{73.6}  & 58.0        & 60.4         & 72.5          & 72.2             & 90.2               & 77.2           & -              \\
\rowcolor[HTML]{E0FFE0} 
infly/INF-ORM-Llama3.1-70B               & \raisebox{-1pt}{\classifier\hspace*{-0.15em}} Classifier                      & \textbf{72.7}  & 60.2        & 77.0         & 76.4          & 79.7             & 69.2               & 76.5           & 95.1           \\
\rowcolor[HTML]{E0FFE0} 
Skywork/Skywork-Reward-V2-Llama-3.1-8B   & \raisebox{-1pt}{\classifier\hspace*{-0.15em}} Classifier                      & \textbf{71.5}  & 85.9        & 57.7         & 74.9          & 74.5             & 71.7               & 74.5           & -              \\
\rowcolor[HTML]{E0FFE0} 
allenai/Llama-3.1-8B-Instruct-RM-RB2     & \raisebox{-1pt}{\classifier\hspace*{-0.15em}} Classifier                      & \textbf{71.5}  & 54.0        & 74.5         & 72.5          & 78.8             & 72.9               & 72.8           & 88.8           \\
\rowcolor[HTML]{E0FFE0} 
Skywork/Skywork-VL-Reward-7B             & \raisebox{-1pt}{\classifier\hspace*{-0.15em}} Classifier                      & \textbf{71.4}  & 59.3        & 74.5         & 73.8          & 74.1             & 72.3               & 68.8           & 90.1           \\
\rowcolor[HTML]{E0FFE0} 
allenai/Llama-3.1-Tulu-3-8B-DPO-RM-RB2   & \raisebox{-1pt}{\classifier\hspace*{-0.15em}} Classifier                      & \textbf{70.2}  & 61.1        & 73.2         & 73.2          & 76.6             & 67.9               & 68.7           & 84.3           \\
\rowcolor[HTML]{E0FFE0} 
allenai/Llama-3.1-Tulu-3-8B-RL-RM-RB2    & \raisebox{-1pt}{\classifier\hspace*{-0.15em}} Classifier                      & \textbf{70.0}  & 55.3        & 71.9         & 74.1          & 76.9             & 68.6               & 68.7           & 83.7           \\
\rowcolor[HTML]{E0FFE0} 
Skywork/Skywork-Reward-Gemma-2-27B-v0.2  & \raisebox{-1pt}{\classifier\hspace*{-0.15em}} Classifier                      & \textbf{69.9}  & 59.3        & 75.3         & 73.5          & 72.2             & 68.1               & 75.3           & 94.3           \\
\rowcolor[HTML]{FFFFCC } 
anthropic/claude-3-7-sonnet-20250219     & \raisebox{-1.2pt}{\generative\hspace*{-0.1em}} Generative                      & \textbf{69.2}  & 57.1        & 67.2         & 70.9          & 71.9             & 73.4               & 75.4           & -              \\
\rowcolor[HTML]{FFFFCC } 
anthropic/claude-3-5-sonnet-20240620     & \raisebox{-1.2pt}{\generative\hspace*{-0.1em}} Generative                      & \textbf{69.0}  & 57.1        & 69.4         & 72.2          & 72.2             & 70.3               & 64.7           & 84.2           \\
\rowcolor[HTML]{E0FFE0} 
ShikaiChen/LDL-Reward-Gemma-2-27B-v0.1   & \raisebox{-1pt}{\classifier\hspace*{-0.15em}} Classifier                      & \textbf{68.4}  & 59.3        & 74.5         & 71.6          & 71.6             & 65.3               & 72.5           & 95.0           \\
\rowcolor[HTML]{E0FFE0} 
Skywork/Skywork-Reward-Llama-3.1-8B-v0.2 & \raisebox{-1pt}{\classifier\hspace*{-0.15em}} Classifier                      & \textbf{68.2}  & 55.3        & 73.2         & 67.7          & 73.8             & 68.6               & 71.8           & 93.1           \\
\rowcolor[HTML]{E0FFE0} 
allenai/Llama-3.1-8B-Base-RM-RB2         & \raisebox{-1pt}{\classifier\hspace*{-0.15em}} Classifier                      & \textbf{68.1}  & 57.1        & 70.2         & 66.1          & 70.6             & 72.1               & 64.9           & 84.6           \\
\rowcolor[HTML]{E0FFE0} 
Ray2333/GRM-Llama3-8B-rewardmodel-ft     & \raisebox{-1pt}{\classifier\hspace*{-0.15em}} Classifier                      & \textbf{66.6}  & 58.0        & 73.2         & 67.4          & 65.3             & 67.9               & 67.7           & 91.5           \\
\rowcolor[HTML]{E0FFE0} 
allenai/Llama-3.1-Tulu-3-8B-SFT-RM-RB2   & \raisebox{-1pt}{\classifier\hspace*{-0.15em}} Classifier                      & \textbf{66.5}  & 50.0        & 65.5         & 69.0          & 71.9             & 69.9               & 68.2           & 85.5           \\
\rowcolor[HTML]{E0FFE0} 
LxzGordon/URM-LLaMa-3.1-8B               & \raisebox{-1pt}{\classifier\hspace*{-0.15em}} Classifier                      & \textbf{65.4}  & 55.8        & 69.8         & 62.3          & 66.6             & 69.4               & 73.9           & 92.9           \\
\rowcolor[HTML]{FFFFCC } 
openai/gpt-4o-2024-08-06                 & \raisebox{-1.2pt}{\generative\hspace*{-0.1em}} Generative                      & \textbf{60.9}  & 46.5        & 59.6         & 59.1          & 56.9             & 73.1               & 64.9           & 86.7           \\ \hline
\end{tabular}}
\caption{Top performing models on Long-form RewardBench, categorized into two types: classifier (\raisebox{-2pt}{\classifier\hspace*{-0.1em}}) and generative (\raisebox{-2pt}{\generative\hspace*{-0.1em}}). We also present their performance on RewardBench2 (RB2) and RewardBench1 (RB1) as extra reference.}
\label{tab:main}
\end{table*}

\begin{table}[!t]
\resizebox{\linewidth}{!}{
\begin{tabular}{cccc}
\hline
\textbf{Model}                               & \textbf{Type} & \textbf{Score} & \textbf{Invalid} \\ \hline
\multirow{2}{*}{RISE-Judge-Qwen2.5-32B}      & Scoring       & 37.1           & 46.7\%           \\
                                             & Selection     & 51.6           & 23.3\%           \\ \hline
\multirow{2}{*}{Selene-1-Mini-Llama-3.1-8B}  & Scoring       & 25.6           & 71.2\%           \\
                                             & Selection     & 57.3           & 7.9\%            \\ \hline
\multirow{2}{*}{Skywork-Critic-Llama-3.1-8B} & Scoring       & 24.3           & 73.9\%           \\
                                             & Selection     & 25.1           & 68.8\%           \\ \hline
\end{tabular}}
\caption{Fine-tuned generative models used for BoN sampling. We report both the overall score and the percentage of invalid responses on the four BoN sampling subsets.}
\label{tab:fine-tuned}
\end{table}

\begin{table*}[]
\centering
\resizebox{0.8\textwidth}{!}{
\begin{tabular}{cc|cccccc}
\hline
\multirow{2}{*}{\textbf{Model}}                        & \multirow{2}{*}{\textbf{Type}} & \multicolumn{6}{c}{\textbf{Long-form RewardBench}}                                                    \\
                                                       &                                & \textbf{Score} & \textbf{QA} & \textbf{RAG} & \textbf{Chat} & \textbf{Writing} & \textbf{Reasoning} \\ \hline
\multirow{2}{*}{google/gemini-2.5-flash-preview-04-17} & Scoring                        & 40.8           & 20.3        & 19.3         & 41.3          & 36.9             & 64.9               \\
                                                       & Selection                      & \textbf{73.6}  & 58.0        & 60.4         & 72.5          & 72.2             & 90.2               \\ \hline
\multirow{2}{*}{anthropic/claude-opus-4-20250514}      & Scoring                        & 58.9           & 66.4        & 71.8         & 82.5          & 80.0             & 66.4               \\
                                                       & Selection                      & \textbf{74.4}  & 57.1        & 71.9         & 75.4          & 79.4             & 80.1               \\ \hline
\multirow{2}{*}{openai/gpt-4o-2024-08-06}              & Scoring                        & 41.1           & 14.2        & 25.1         & 40.6          & 40.3             & 64.0               \\
                                                       & Selection                      & \textbf{60.9}  & 46.5        & 59.6         & 59.1          & 56.9             & 73.1               \\ \hline
\multirow{2}{*}{anthropic/claude-3-7-sonnet-20250219}  & Scoring                        & 36.7           & 9.3         & 22.6         & 35.5          & 40.9             & 55.7               \\
                                                       & Selection                      & \textbf{69.2}  & 57.1        & 67.2         & 70.9          & 71.9             & 73.4               \\ \hline
\multirow{2}{*}{openai/gpt-4.1-2025-04-14}             & Scoring                        & 53.1           & 36.7        & 34.9         & 56.2          & 62.5             & 62.0               \\
                                                       & Selection                      & \textbf{74.6}  & 58.0        & 70.6         & 80.5          & 78.1             & 78.4               \\ \hline
\multirow{2}{*}{Anthropic/claude-3-5-sonnet-20240620}  & Scoring                        & 29.2           & 19.0        & 6.8          & 21.1          & 25.0             & 54.6               \\
                                                       & Selection                      & \textbf{69.0}  & 57.1        & 69.4         & 72.2          & 72.2             & 70.3               \\ \hline
\end{tabular}}
\caption{Comparison of generative models in Selection or Scoring modes.}
\label{tab:generative}
\end{table*}

\subsection{Experimental Analysis}

Table \ref{tab:main} presents the top-performing results on Long-form RewardBench, revealing several key findings:

\vspace{-2mm}

\paragraph{Finding 1: Sequential Classifiers Perform Better Overall.} Despite their limited size, classifiers consistently perform well, dominating leaderboards and frequently securing top ranks. This aligns with other reward benchmarks like RewardBench, where most top-performing reward models are sequential classifiers. Conversely, many powerful generative models, despite their superior performance in other tasks like question answering, only achieve intermediate results across most subsets. This indicates that these generative models currently underperform in preference modeling, partly due to a lack of relevant data in their training.

\vspace{-2mm}

\paragraph{Finding 2: Generative Models Outperform on Reasoning Subset.} While most classifiers achieve only around 70\% accuracy on math-reasoning tasks, generative models demonstrate superior performance, reaching over 90\% accuracy. This highlights the critical role of chain-of-thought for evaluating reasoning tasks. Lacking this, classifiers can only rely on statistical features, which inherently limits their accuracy to approximately 70\%.

\vspace{-2mm}

\paragraph{Finding 3: Generative Models Degrade in Scoring Mode.} As explained in Section 3.3, we evaluate generative models in our benchmark in two modes: Scoring and Selection. However, the Scoring-based method consistently yields significantly lower performance compared to Selection, as detailed in Table \ref{tab:generative}. This is because generative models, when used for scoring, frequently assign identical scores to different responses due to their similar quality. In our evaluation, if a rejected response receives the same score as a chosen response, the entire sample is considered incorrect. In contrast, the sequential classifier does not have such concern as it would always assign different scores to different responses. Consequently, all reported results of generative models in Table \ref{tab:main} are based on Selection mode.

\vspace{-2mm}


\paragraph{Finding 4: Fine-tuned Generative Models Struggle with BoN Sampling.} Fine-tuned local generative judge models are proposed to reduce reliance on external APIs and computational overhead. Following RewardBench1, we also attempted to evaluate them on Long-form RewardBench. However, as shown in Table \ref{tab:fine-tuned}, most fine-tuned generative models are overfitted to pairwise selection. When applied for either best-of-n selection or pointwise scoring, these models largely fail to follow instructions for generating answers in the valid format. Therefore, these fine-tuned generative judge models can hardly be used for best-of-n sampling.

%% file: content/5_Analysis.tex
\begin{figure*}[!ht] 
    \centering 

    \begin{subfigure}[b]{0.47\textwidth}
        \centering
        \includegraphics[width=1.0\linewidth]{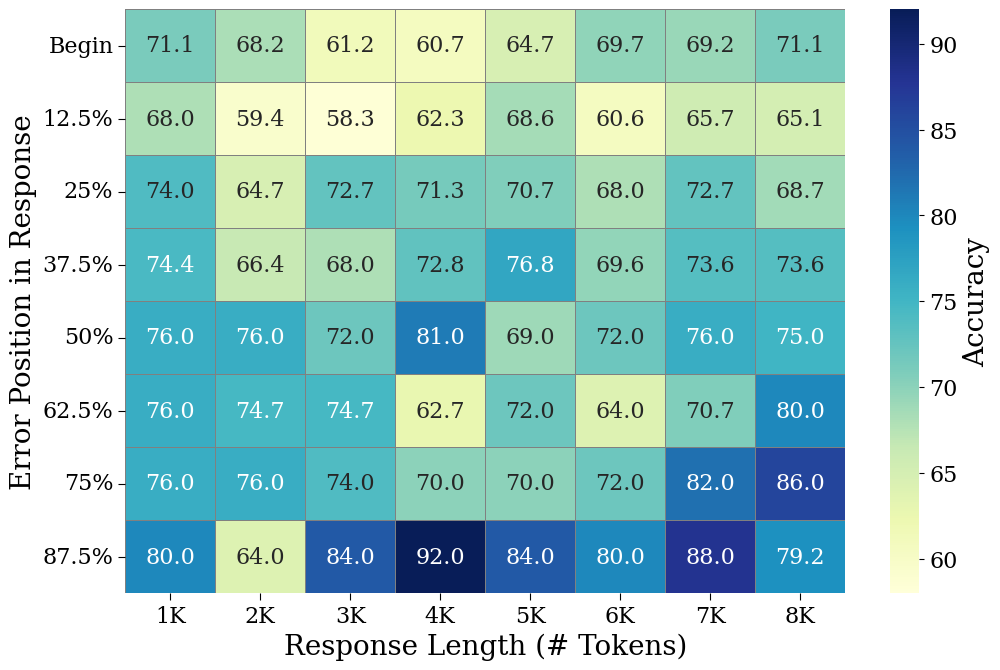}
        \caption{Skywork-Llama-3.1-8B-v0.2 on Safety dimension.}
        \label{fig:generative-safety}
    \end{subfigure}
    \hfill
    \begin{subfigure}[b]{0.47\textwidth}
        \centering
        \includegraphics[width=1.0\linewidth]{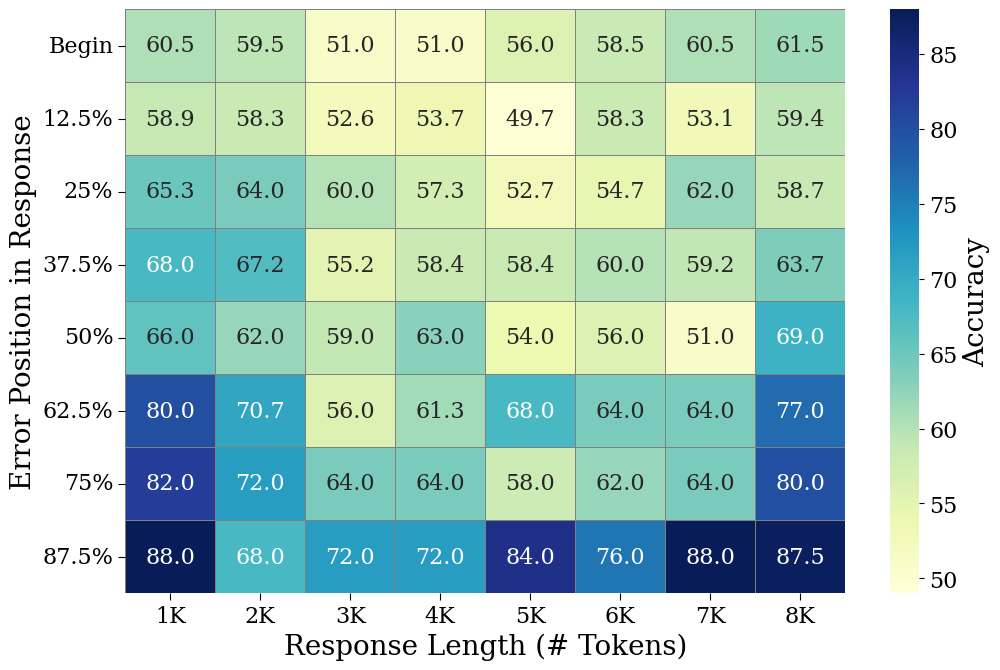}
        \caption{Skywork-Llama-3.1-8B-v0.2 on Factuality dimension.}
        \label{fig:generative-factuality}
    \end{subfigure}

    \vspace{1em} 

    \begin{subfigure}[b]{0.47\textwidth} 
        \centering
        \includegraphics[width=1.0\linewidth]{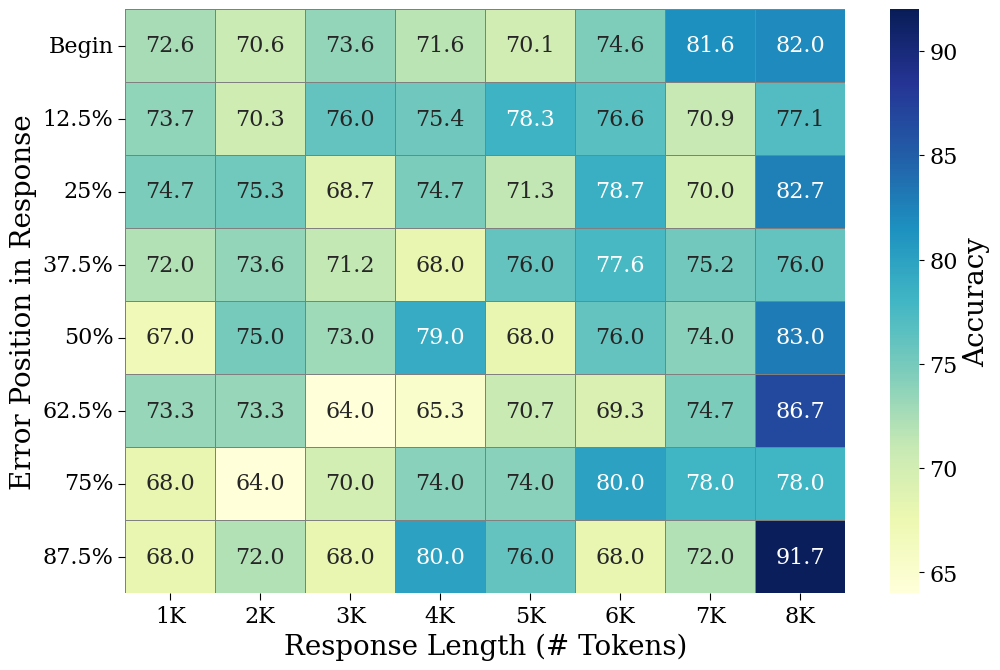}
        \caption{GPT-4o-0806 on Safety dimension.}
        \label{fig:classifier-safety}
    \end{subfigure}
    \hfill 
    \begin{subfigure}[b]{0.47\textwidth}
        \centering
        \includegraphics[width=1.0\linewidth]{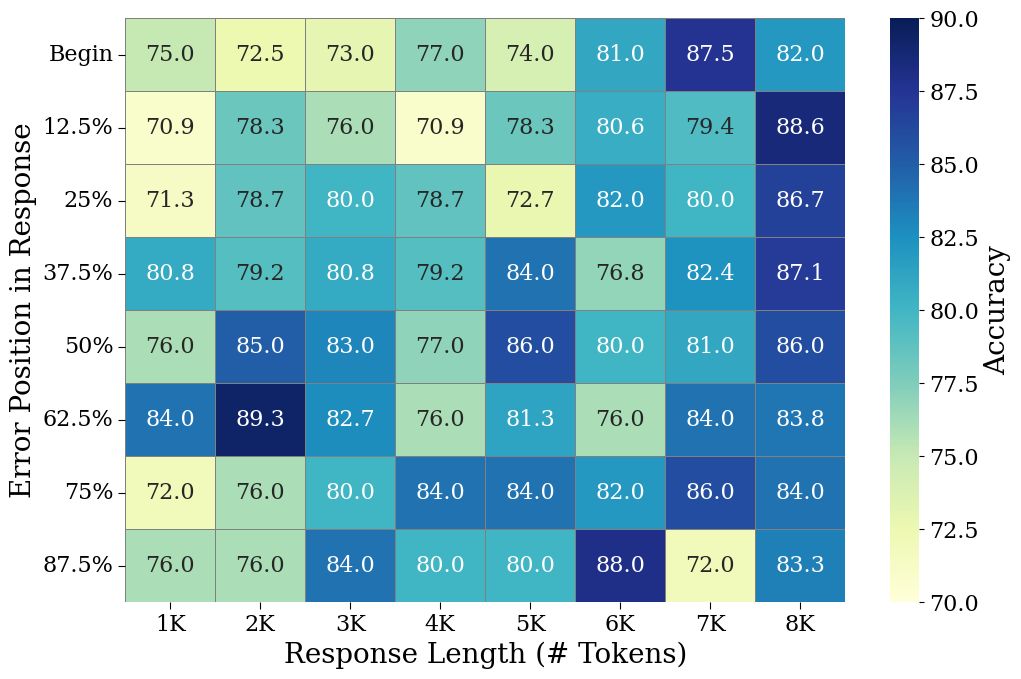}
        \caption{GPT-4o-0806 on Factuality dimension.}
        \label{fig:classifier-factuality}
    \end{subfigure}

    \caption{Needle-in-a-haystack test of two representative reward models on Factuality and Safety.} 
    \label{fig:needle-in-a-haystack} 
\end{figure*}

\section{Long-form Needle-in-a-Haystack Test}

\subsection{Dataset Construction}

Inspired by the recent needle-in-a-haystack test \cite{schuster2025needle}, we introduce a Long-form Needle-in-a-Haystack Test. Our objective is to evaluate how effectively a reward model can detect a specific error embedded at varying locations within responses of different lengths. For this, we constructed a fine-grained evaluation set with two easily manipulable dimensions: \textit{Factuality} and \textit{Safety}. To exclude confounding variables, we meticulously designed the test set construction process as follows:

\begin{enumerate}
    \item Randomly selected 100 instructions from Longform-RewardBench. Notice we did not include instructions from Reasoning Subset, as inserting error to reasoning process would disrupt the logical flow.
    \item For each instruction, we leveraged gemini-2.5 to generate responses of eight different lengths, ranging from 1K to 8K tokens. To ensure responses met the required length, we verified their token count and extended them if necessary until the requirement was satisfied. 
    \item To create rejected samples, we inserted either incorrect factual information or harmful content into the generated responses. To mitigate potential fluency issues and ensure a fair comparison, a similar but factually correct or harmless counterpart was simultaneously inserted at the corresponding position in the chosen responses.
    \item For \textit{Factuality} samples, the information was constructed by gpt-4o based on the immediate context of the response. For \textit{Safety} samples, predefined harmful phrases, derived from common safety issues reported in \cite{zhang2024safetybenchevaluatingsafetylarge}, were utilized.
    \item These insertions were uniformly distributed throughout the response. The number of insertions ($n_{insert}$) was scaled with the response length ($l_{resp}$) by the formula $n_{insert} = l_{resp}/1000$. This approach ensures a consistent density of inserted content across varying lengths.
\end{enumerate}

These steps ensure that across different sample groups, the error type and response content remain consistent, with only error position and response length varying\footnote{Generating responses exceeding 1K tokens remains a challenge for most current LLMs due to their limited instruction-following ability. This limitation is also reported in \cite{que2024hellobench}. For this reason, gemini-2.5 was chosen, given its demonstrated proficiency in generating responses beyond 2K tokens.}. Following this process, a total of 7,200 sample pairs were generated across both dimensions. Each error (referred to as the ``needle") was randomly and uniformly inserted into the response's content (referred to as the ``haystack").

\subsection{Experimental Analysis}
We select two representative models for needle-in-the-haystack evaluation, one sequence classifier and one generative model, namely Skywork-Llama-3.1-8B-v0.2 and gpt-4o-0806. We apply the models to both testsets, and calculated the accuracy within each group, as shown in Figure \ref{fig:needle-in-a-haystack}.

For the generative model, \textbf{a significant correlation was observed between its error detection accuracy and both response length and error position.} Specifically, accuracy was relatively high for shorter responses (e.g., 1K-2K tokens) but generally declined with increasing response length. Furthermore, accuracy was lower when errors were situated in the middle of a response, whereas it was higher when errors appeared at the beginning or end. This behavior is consistent with the ``lost-in-the-middle" phenomenon previously observed in long-context scenarios \cite{liu2023lost}.

In contrast, \textbf{the sequence classifier presents greater robustness concerning response length compared to the generative model.} However, the classifier exhibited a strong correlation with error position, achieving higher accuracy when errors were located towards the end of the document. This could be attributed to the typical training paradigm for classification models on preference pairs, where the pairs often differs predominantly in the latter part of the response (as the former part is the instruction). Consequently, during prediction, the reward model's attention tends to be largely focused on the latter half of the input.

\begin{figure}[!t]
    \centering
    \begin{subfigure}[b]{0.48\textwidth}
        \hspace*{-0.1in}
        \centering
        \includegraphics[width=\textwidth]{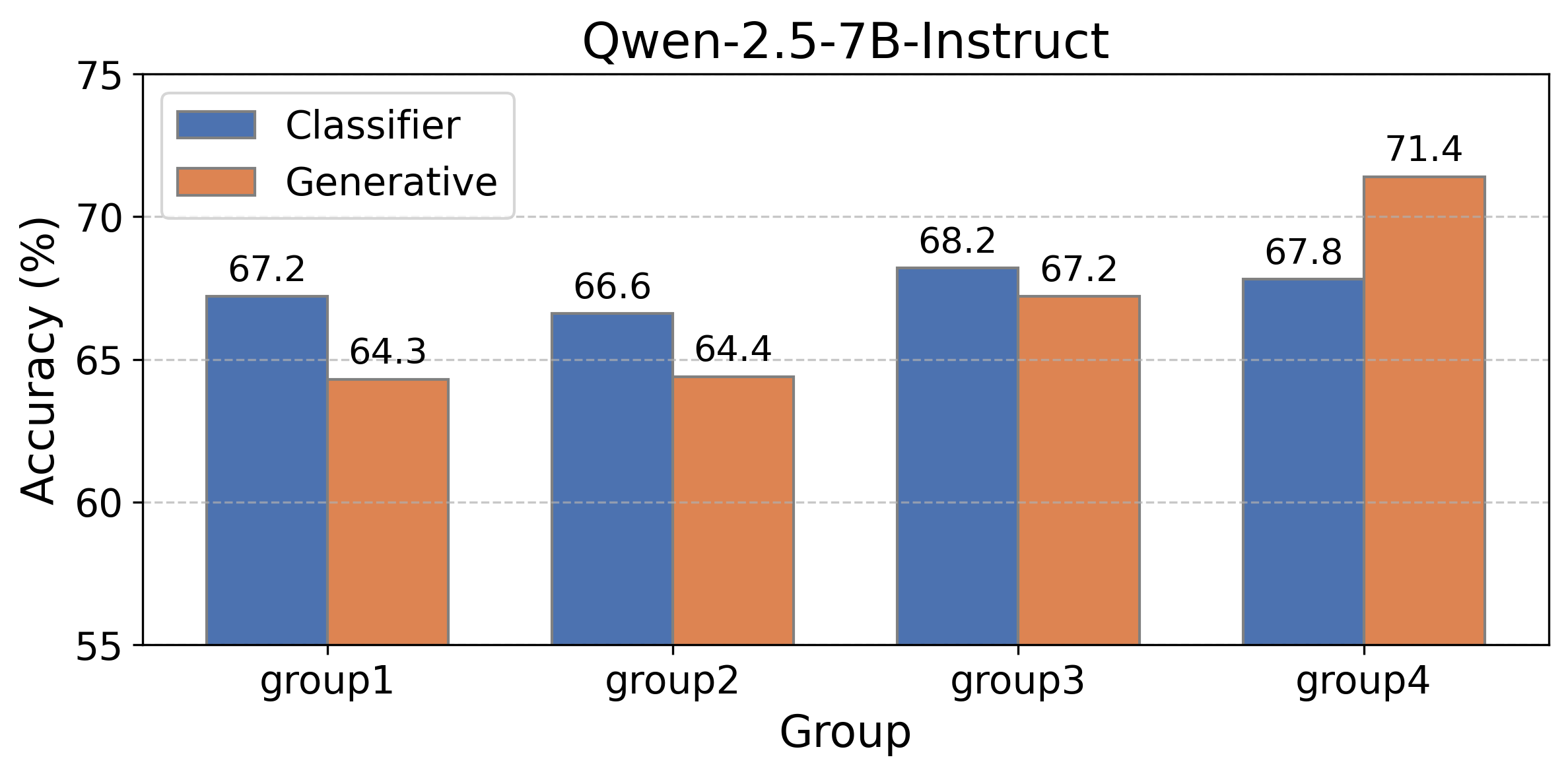} 
        \label{fig:subim1}
    \end{subfigure}
    \begin{subfigure}[b]{0.48\textwidth}
        \hspace*{-0.1in}
        \centering
        \includegraphics[width=\textwidth]{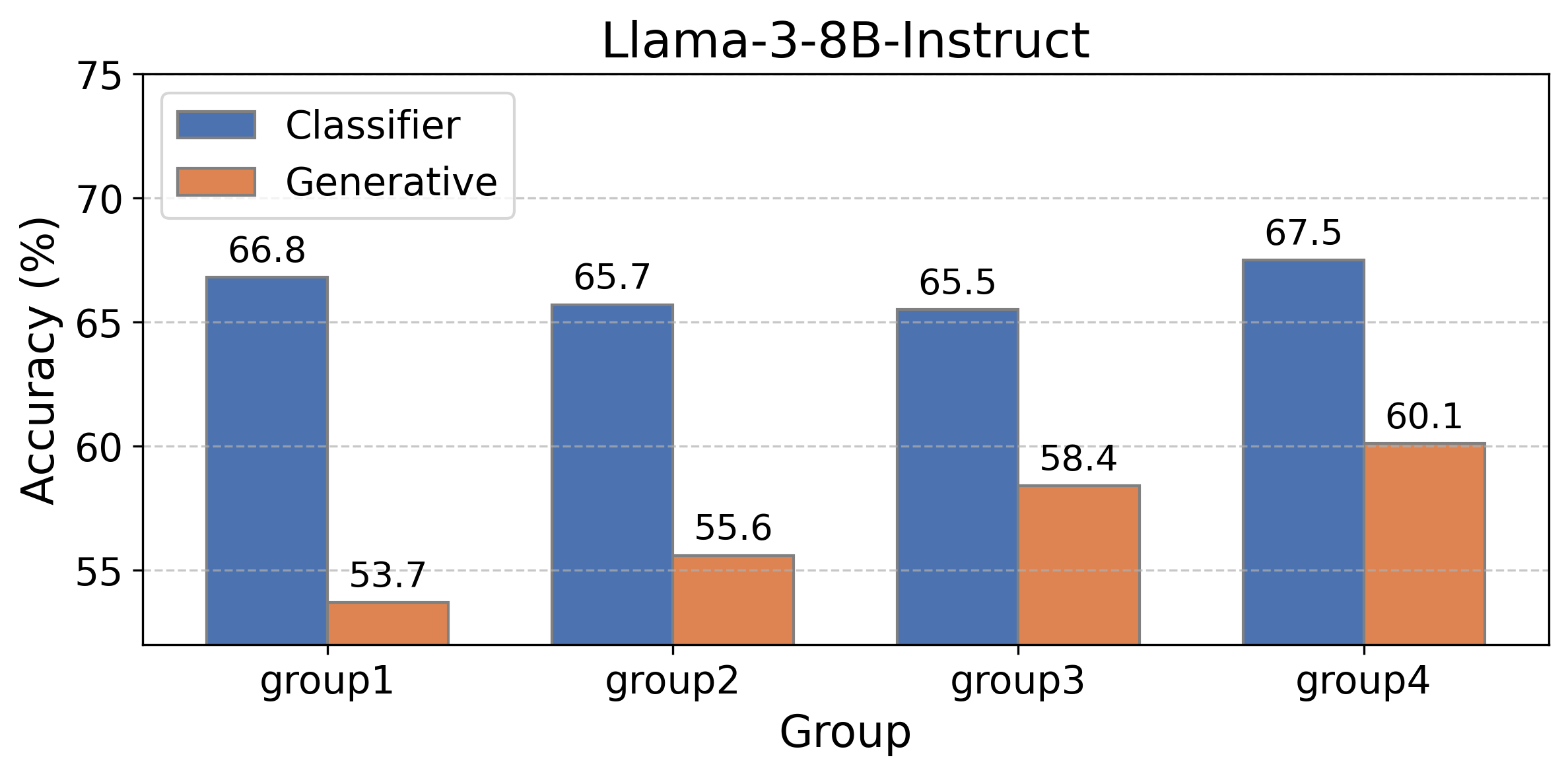} 
        \label{fig:subim2}
    \end{subfigure} 
    \caption{RM accuracy with different training data lengths.}
    \label{fig:perform-dist}
\end{figure}


\subsection{Does Training Length Matter?}

In this section, we investigate the influence of training length on reward model performance. While Long-form RewardBench targets the reward modeling of long-form responses, it is unclear if the training data itself needs to reflect this extended length. To explore this, we constructed a training set based on the pipeline depicted in Section 3.1. We first generated instructions using the instruction back-translation technique described in \cite{li2023self}. Following this, we applied the pipeline to construct responses, preferences, and annotations, ultimately deriving 30k preference pairs. Subsequently, we divided these responses into four groups based on length. We then fine-tuned two types of reward models: sequence classifiers and generative models, and then evaluated them on Long-form RewardBench.

As shown in Figure \ref{fig:perform-dist}, \textbf{classifiers demonstrate insensitivity to training length}, showing stable performance with minimal fluctuation across all groups. In contrast, \textbf{generative models exhibit clear sensitivity to training length}, tending to achieve higher accuracy in longer text groups. This suggests that classifiers can scale effectively to longer sequences even when not specifically trained on them. This finding is consistent with the results presented in Section 5.2, where only the performance of generative models showed a strong correlation with response length. 

Furthermore, we also notice that compared with classifiers, the performance of generative models is highly dependent on the base model, with Qwen-based generative RMs significantly outperforms Llama-based generative RMs.



%% file: content/6_Conclusion.tex
\section{Conclusion}

In this paper, we introduce Long-form RewardBench, a novel benchmark designed to evaluate the preference modeling capabilities of reward models for long-form text generation. Our findings reveal that despite their high overall performance, current reward models significantly lag when specifically evaluated on long-form reward modeling. We further show that different reward models exhibit distinct trends regarding error position and response length. We believe Long-form RewardBench will guide developers and researchers in understanding the long-form preference modeling capabilities of reward models and ultimately facilitate advancements in aligning long-text generation scenarios.

\section*{Acknowledgements}
This work is supported by National Natural Science Foundation of China (62276077, 62376075, 62376076), Department of Science and Technology of Heilongjiang (Grant No. ZY04JD04) , Jiangsu Science and Technology Major Project (BG2024031) and Nanjing University AI \& AI for Science Funding (2024300540).

%% file: aaai2026.bib
@inproceedings{lambert2025rewardbench,
  title={RewardBench: Evaluating Reward Models for Language Modeling},
  author={Lambert, Nathan and Pyatkin, Valentina and Morrison, Jacob and Miranda, Lester James Validad and Lin, Bill Yuchen and Chandu, Khyathi and Dziri, Nouha and Kumar, Sachin and Zick, Tom and Choi, Yejin and others},
  booktitle={Findings of the Association for Computational Linguistics: NAACL 2025},
  pages={1755--1797},
  year={2025}
}

@article{zhong2025comprehensive,
  title={A comprehensive survey of reward models: Taxonomy, applications, challenges, and future},
  author={Zhong, Jialun and Shen, Wei and Li, Yanzeng and Gao, Songyang and Lu, Hua and Chen, Yicheng and Zhang, Yang and Zhou, Wei and Gu, Jinjie and Zou, Lei},
  journal={arXiv preprint arXiv:2504.12328},
  year={2025}
}

@misc{dong2024rlhfworkflowrewardmodeling,
  title={RLHF Workflow: From Reward Modeling to Online RLHF}, 
  author={Hanze Dong and Wei Xiong and Bo Pang and Haoxiang Wang and Han Zhao and Yingbo Zhou and Nan Jiang and Doyen Sahoo and Caiming Xiong and Tong Zhang},
  year={2024},
  eprint={2405.07863},
  archivePrefix={arXiv},
  primaryClass={cs.LG},
  url={https://arxiv.org/abs/2405.07863}, 
}

@article{zhang2024prototypical,
  title={Prototypical reward network for data-efficient rlhf},
  author={Zhang, Jinghan and Wang, Xiting and Jin, Yiqiao and Chen, Changyu and Zhang, Xinhao and Liu, Kunpeng},
  journal={arXiv preprint arXiv:2406.06606},
  year={2024}
}

@article{huang2025think,
  title={Think-j: Learning to think for generative llm-as-a-judge},
  author={Huang, Hui and He, Yancheng and Zhou, Hongli and Zhang, Rui and Liu, Wei and Wang, Weixun and Su, Wenbo and Zheng, Bo and Liu, Jiaheng},
  journal={arXiv preprint arXiv:2505.14268},
  year={2025}
}

@article{he2025can,
  title={Can large language models detect errors in long chain-of-thought reasoning?},
  author={He, Yancheng and Li, Shilong and Liu, Jiaheng and Wang, Weixun and Bu, Xingyuan and Zhang, Ge and Peng, Zhongyuan and Zhang, Zhaoxiang and Zheng, Zhicheng and Su, Wenbo and others},
  journal={arXiv preprint arXiv:2502.19361},
  year={2025}
}

@article{malik2025rewardbench,
  title={RewardBench 2: Advancing Reward Model Evaluation},
  author={Malik, Saumya and Pyatkin, Valentina and Land, Sander and Morrison, Jacob and Smith, Noah A and Hajishirzi, Hannaneh and Lambert, Nathan},
  journal={arXiv preprint arXiv:2506.01937},
  year={2025}
}

@article{que2024hellobench,
  title={Hellobench: Evaluating long text generation capabilities of large language models},
  author={Que, Haoran and Duan, Feiyu and He, Liqun and Mou, Yutao and Zhou, Wangchunshu and Liu, Jiaheng and Rong, Wenge and Wang, Zekun Moore and Yang, Jian and Zhang, Ge and others},
  journal={arXiv preprint arXiv:2409.16191},
  year={2024}
}

@article{zhou2024rmb,
  title={RMB: Comprehensively benchmarking reward models in LLM alignment},
  author={Zhou, Enyu and Zheng, Guodong and Wang, Binghai and Xi, Zhiheng and Dou, Shihan and Bao, Rong and Shen, Wei and Xiong, Limao and Fan, Jessica and Mou, Yurong and others},
  journal={arXiv preprint arXiv:2410.09893},
  year={2024}
}

@article{frick2024evaluate,
  title={How to evaluate reward models for rlhf},
  author={Frick, Evan and Li, Tianle and Chen, Connor and Chiang, Wei-Lin and Angelopoulos, Anastasios N and Jiao, Jiantao and Zhu, Banghua and Gonzalez, Joseph E and Stoica, Ion},
  journal={arXiv preprint arXiv:2410.14872},
  year={2024}
}

@article{liu2024rm,
  title={Rm-bench: Benchmarking reward models of language models with subtlety and style},
  author={Liu, Yantao and Yao, Zijun and Min, Rui and Cao, Yixin and Hou, Lei and Li, Juanzi},
  journal={arXiv preprint arXiv:2410.16184},
  year={2024}
}

@article{gureja2024m,
  title={M-RewardBench: Evaluating reward models in multilingual settings},
  author={Gureja, Srishti and Miranda, Lester James V and Islam, Shayekh Bin and Maheshwary, Rishabh and Sharma, Drishti and Winata, Gusti and Lambert, Nathan and Ruder, Sebastian and Hooker, Sara and Fadaee, Marzieh},
  journal={arXiv preprint arXiv:2410.15522},
  year={2024}
}

@misc{yasunaga2025multimodalrewardbenchholisticevaluation,
      title={Multimodal RewardBench: Holistic Evaluation of Reward Models for Vision Language Models}, 
      author={Michihiro Yasunaga and Luke Zettlemoyer and Marjan Ghazvininejad},
      year={2025},
      eprint={2502.14191},
      archivePrefix={arXiv},
      primaryClass={cs.CV},
      url={https://arxiv.org/abs/2502.14191}, 
}

@article{jin2024rag,
  title={Rag-rewardbench: Benchmarking reward models in retrieval augmented generation for preference alignment},
  author={Jin, Zhuoran and Yuan, Hongbang and Men, Tianyi and Cao, Pengfei and Chen, Yubo and Liu, Kang and Zhao, Jun},
  journal={arXiv preprint arXiv:2412.13746},
  year={2024}
}

@article{kim2024evaluating,
  title={Evaluating robustness of reward models for mathematical reasoning},
  author={Kim, Sunghwan and Kang, Dongjin and Kwon, Taeyoon and Chae, Hyungjoo and Won, Jungsoo and Lee, Dongha and Yeo, Jinyoung},
  journal={arXiv preprint arXiv:2410.01729},
  year={2024}
}

@article{wu2024longgenbench,
  title={Longgenbench: Benchmarking long-form generation in long context llms},
  author={Wu, Yuhao and Hee, Ming Shan and Hu, Zhiqing and Lee, Roy Ka-Wei},
  journal={arXiv preprint arXiv:2409.02076},
  year={2024}
}

@article{zhao2024wildchat,
  title={Wildchat: 1m chatgpt interaction logs in the wild},
  author={Zhao, Wenting and Ren, Xiang and Hessel, Jack and Cardie, Claire and Choi, Yejin and Deng, Yuntian},
  journal={arXiv preprint arXiv:2405.01470},
  year={2024}
}

@article{bai2024longwriter,
  title={Longwriter: Unleashing 10,000+ word generation from long context llms},
  author={Bai, Yushi and Zhang, Jiajie and Lv, Xin and Zheng, Linzhi and Zhu, Siqi and Hou, Lei and Dong, Yuxiao and Tang, Jie and Li, Juanzi},
  journal={arXiv preprint arXiv:2408.07055},
  year={2024}
}

@misc{gao2024omnimathuniversalolympiadlevel,
      title={Omni-MATH: A Universal Olympiad Level Mathematic Benchmark For Large Language Models}, 
      author={Bofei Gao and Feifan Song and Zhe Yang and Zefan Cai and Yibo Miao and Qingxiu Dong and Lei Li and Chenghao Ma and Liang Chen and Runxin Xu and Zhengyang Tang and Benyou Wang and Daoguang Zan and Shanghaoran Quan and Ge Zhang and Lei Sha and Yichang Zhang and Xuancheng Ren and Tianyu Liu and Baobao Chang},
      year={2024},
      eprint={2410.07985},
      archivePrefix={arXiv},
      primaryClass={cs.CL},
      url={https://arxiv.org/abs/2410.07985}, 
}

@article{he2024olympiadbench,
  title={Olympiadbench: A challenging benchmark for promoting agi with olympiad-level bilingual multimodal scientific problems},
  author={He, Chaoqun and Luo, Renjie and Bai, Yuzhuo and Hu, Shengding and Thai, Zhen Leng and Shen, Junhao and Hu, Jinyi and Han, Xu and Huang, Yujie and Zhang, Yuxiang and others},
  journal={arXiv preprint arXiv:2402.14008},
  year={2024}
}

@article{hendrycks2021measuring,
  title={Measuring mathematical problem solving with the math dataset},
  author={Hendrycks, Dan and Burns, Collin and Kadavath, Saurav and Arora, Akul and Basart, Steven and Tang, Eric and Song, Dawn and Steinhardt, Jacob},
  journal={arXiv preprint arXiv:2103.03874},
  year={2021}
}

@inproceedings{schuster2025needle,
  title={Needle-in-the-Haystack Testing LLMs with a Complex Reasoning Task},
  author={Schuster, Thomas and Lambert, Marian and D{\"o}ring, Nico and Tr{\"o}gele, Julius},
  booktitle={International Conference on Engineering Applications of Neural Networks},
  pages={254--266},
  year={2025},
  organization={Springer}
}

@misc{zhang2024safetybenchevaluatingsafetylarge,
      title={SafetyBench: Evaluating the Safety of Large Language Models}, 
      author={Zhexin Zhang and Leqi Lei and Lindong Wu and Rui Sun and Yongkang Huang and Chong Long and Xiao Liu and Xuanyu Lei and Jie Tang and Minlie Huang},
      year={2024},
      eprint={2309.07045},
      archivePrefix={arXiv},
      primaryClass={cs.CL},
      url={https://arxiv.org/abs/2309.07045}, 
}

@article{liu2023lost,
  title={Lost in the middle: How language models use long contexts},
  author={Liu, Nelson F and Lin, Kevin and Hewitt, John and Paranjape, Ashwin and Bevilacqua, Michele and Petroni, Fabio and Liang, Percy},
  journal={arXiv preprint arXiv:2307.03172},
  year={2023}
}

@article{li2023self,
  title={Self-alignment with instruction backtranslation},
  author={Li, Xian and Yu, Ping and Zhou, Chunting and Schick, Timo and Levy, Omer and Zettlemoyer, Luke and Weston, Jason and Lewis, Mike},
  journal={arXiv preprint arXiv:2308.06259},
  year={2023}
}

@article{wu2025longwriter,
  title={LongWriter-Zero: Mastering Ultra-Long Text Generation via Reinforcement Learning},
  author={Wu, Yuhao and Bai, Yushi and Hu, Zhiqiang and Lee, Roy Ka-Wei and Li, Juanzi},
  journal={arXiv preprint arXiv:2506.18841},
  year={2025}
}

@article{team2024qwen2,
  title={Qwen2 technical report},
  author={Team, Qwen},
  journal={arXiv preprint arXiv:2407.10671},
  year={2024}
}

@article{grattafiori2024llama,
  title={The llama 3 herd of models},
  author={Grattafiori, Aaron and Dubey, Abhimanyu and Jauhri, Abhinav and Pandey, Abhinav and Kadian, Abhishek and Al-Dahle, Ahmad and Letman, Aiesha and Mathur, Akhil and Schelten, Alan and Vaughan, Alex and others},
  journal={arXiv preprint arXiv:2407.21783},
  year={2024}
}

@misc{jiang2023mistral7b,
      title={Mistral 7B}, 
      author={Albert Q. Jiang and Alexandre Sablayrolles and Arthur Mensch and Chris Bamford and Devendra Singh Chaplot and Diego de las Casas and Florian Bressand and Gianna Lengyel and Guillaume Lample and Lucile Saulnier and Lélio Renard Lavaud and Marie-Anne Lachaux and Pierre Stock and Teven Le Scao and Thibaut Lavril and Thomas Wang and Timothée Lacroix and William El Sayed},
      year={2023},
      eprint={2310.06825},
      archivePrefix={arXiv},
      primaryClass={cs.CL},
      url={https://arxiv.org/abs/2310.06825}, 
}

@article{liu2024deepseek,
  title={Deepseek-v3 technical report},
  author={Liu, Aixin and Feng, Bei and Xue, Bing and Wang, Bingxuan and Wu, Bochao and Lu, Chengda and Zhao, Chenggang and Deng, Chengqi and Zhang, Chenyu and Ruan, Chong and others},
  journal={arXiv preprint arXiv:2412.19437},
  year={2024}
}

@article{saito2023verbosity,
  title={Verbosity bias in preference labeling by large language models},
  author={Saito, Keita and Wachi, Akifumi and Wataoka, Koki and Akimoto, Youhei},
  journal={arXiv preprint arXiv:2310.10076},
  year={2023}
}

@misc{yu2025improvellmasajudgeabilitygeneral,
      title={Improve LLM-as-a-Judge Ability as a General Ability}, 
      author={Jiachen Yu and Shaoning Sun and Xiaohui Hu and Jiaxu Yan and Kaidong Yu and Xuelong Li},
      year={2025},
      eprint={2502.11689},
      archivePrefix={arXiv},
      primaryClass={cs.CL},
      url={https://arxiv.org/abs/2502.11689}, 
}

@misc{pham2024surimulticonstraintinstructionfollowing,
      title={Suri: Multi-constraint Instruction Following for Long-form Text Generation}, 
      author={Chau Minh Pham and Simeng Sun and Mohit Iyyer},
      year={2024},
      eprint={2406.19371},
      archivePrefix={arXiv},
      primaryClass={cs.CL},
      url={https://arxiv.org/abs/2406.19371}, 
}
